\documentclass[acmsmall]{acmart}
\AtBeginDocument{%
  }

\setcopyright{acmlicensed}
\copyrightyear{2018}
\acmYear{2018}
\acmDOI{XXXXXXX.XXXXXXX}

\acmJournal{JACM}
\acmVolume{37}
\acmNumber{4}
\acmArticle{111}
\acmMonth{8}

\usepackage{multirow}
\begin{document}

%%
%% The "title" command has an optional parameter,
%% allowing the author to define a "short title" to be used in page headers.
\title{MambaVesselNet++: A Hybrid CNN-Mamba Architecture for Medical Image Segmentation}

%%
%% The "author" command and its associated commands are used to define
%% the authors and their affiliations.
%% Of note is the shared affiliation of the first two authors, and the
%% "authornote" and "authornotemark" commands
%% used to denote shared contribution to the research.
\author{Qing Xu}
% \authornote{Both authors contributed equally to this research.}
% Work done when Zhenye was an intern of Tsinghua University.
\email{qing.xu@nottingham.edu.cn
}

\author{Yanming Chen}
\email{yanming.chen@yale.edu}
\author{Yue Li}
\email{yue.li3@nottingham.edu.cn}
\author{Ziyu Liu}
\email{ziyu.liu@nottingham.edu.cn}
\affiliation{%
  \institution{University of Nottingham Ningbo China}
  \city{Ningbo}
  \country{China}
}

\author{Zhenye Lou}
%\authornotemark[1]
\email{leonlou0921@gmail.com}
\affiliation{%
\institution{Sichuan University}
\city{Chengdu}
\country{China}
}

\author{Yixuan Zhang}
\email{scxyz8@nottingham.edu.cn}
\author{Huizhong Zheng}
\email{huizhong.zheng@nottingham.edu.cn}

\affiliation{%
  \institution{University of Nottingham Ningbo China}
  \city{Ningbo}
  \country{China}
}

% \author{Zhen Chen}
% \email{zhen.chen@yale.edu}
% \affiliation{%
%   \institution{Yale University}
%   \city{New Haven}
%   \country{USA}
% }

\author{Xiangjian He}
\authornote{Corresponding Author.}
\email{sean.he@nottingham.edu.cn}
\affiliation{%
  \institution{University of Nottingham Ningbo China}
  \city{Ningbo}
  \country{China}
}

%%
%% By default, the full list of authors will be used in the page
%% headers. Often, this list is too long, and will overlap
%% other information printed in the page headers. This command allows
%% the author to define a more concise list
%% of authors' names for this purpose.
\renewcommand{\shortauthors}{Trovato et al.}

%%
%% The abstract is a short summary of the work to be presented in the
%% article.
\begin{abstract}
 Medical image segmentation plays an important role in computer-aided diagnosis. Traditional convolution-based U-shape segmentation architectures are usually limited by the local receptive field. Existing vision transformers have been widely applied to diverse medical segmentation frameworks due to their superior capabilities of capturing global contexts. Despite the advantage, the real-world application of vision transformers is challenged by their non-linear self-attention mechanism, requiring huge computational costs. To address this issue, the selective state space model (SSM) Mamba has gained recognition for its adeptness in modeling long-range dependencies in sequential data, particularly noted for its efficient memory costs. In this paper, we propose MambaVesselNet++, a Hybrid CNN-Mamba framework for medical image segmentation. Our MambaVesselNet++ is comprised of a hybrid image encoder (Hi-Encoder) and a bifocal fusion decoder (BF-Decoder). In Hi-Encoder, we first devise the texture-aware layer to capture low-level semantic features by leveraging convolutions. Then, we utilize Mamba to effectively model long-range dependencies with linear complexity. The Bi-Decoder adopts skip connections to combine local and global information of the Hi-Encoder for the accurate generation of segmentation masks. Extensive experiments demonstrate that MambaVesselNet++ outperforms current convolution-based, transformer-based, and Mamba-based state-of-the-arts across diverse medical 2D, 3D, and instance segmentation tasks. The code is available at \url{https://github.com/CC0117/MambaVesselNet}.
\end{abstract}

%%
%% The code below is generated by the tool at http://dl.acm.org/ccs.cfm.
%% Please copy and paste the code instead of the example below.
%%


%%
%% Keywords. The author(s) should pick words that accurately describe
%% the work being presented. Separate the keywords with commas.

% \received{20 February 2007}
% \received[revised]{12 March 2009}
% \received[accepted]{5 June 2009}

%%
%% This command processes the author and affiliation and title
%% information and builds the first part of the formatted document.
\maketitle

\section{Introduction}
In the field of digital healthcare, medical image segmentation stands as a cornerstone for delineating disease regions, grading diseases, and planning treatment \cite{azad2024medical, chen2025sam}. Despite its significance, this poses substantial difficulties across different imaging modalities, e.g., the complex 3D nature of the vascular structures, discrete nuclei distribution, and low-contrast tumor region boundaries. Traditionally, such segmentation tasks are heavily relied on by medical experts, leading to time-consuming and labor-intensive procedures \cite{chen2021super, chen2022personalized}. In this challenging background, automatic segmentation methods are expected to accelerate evaluation
time and improve diagnostic efficiency.

\begin{figure}[!t]
    \centering
    \includegraphics[width=0.55\linewidth]{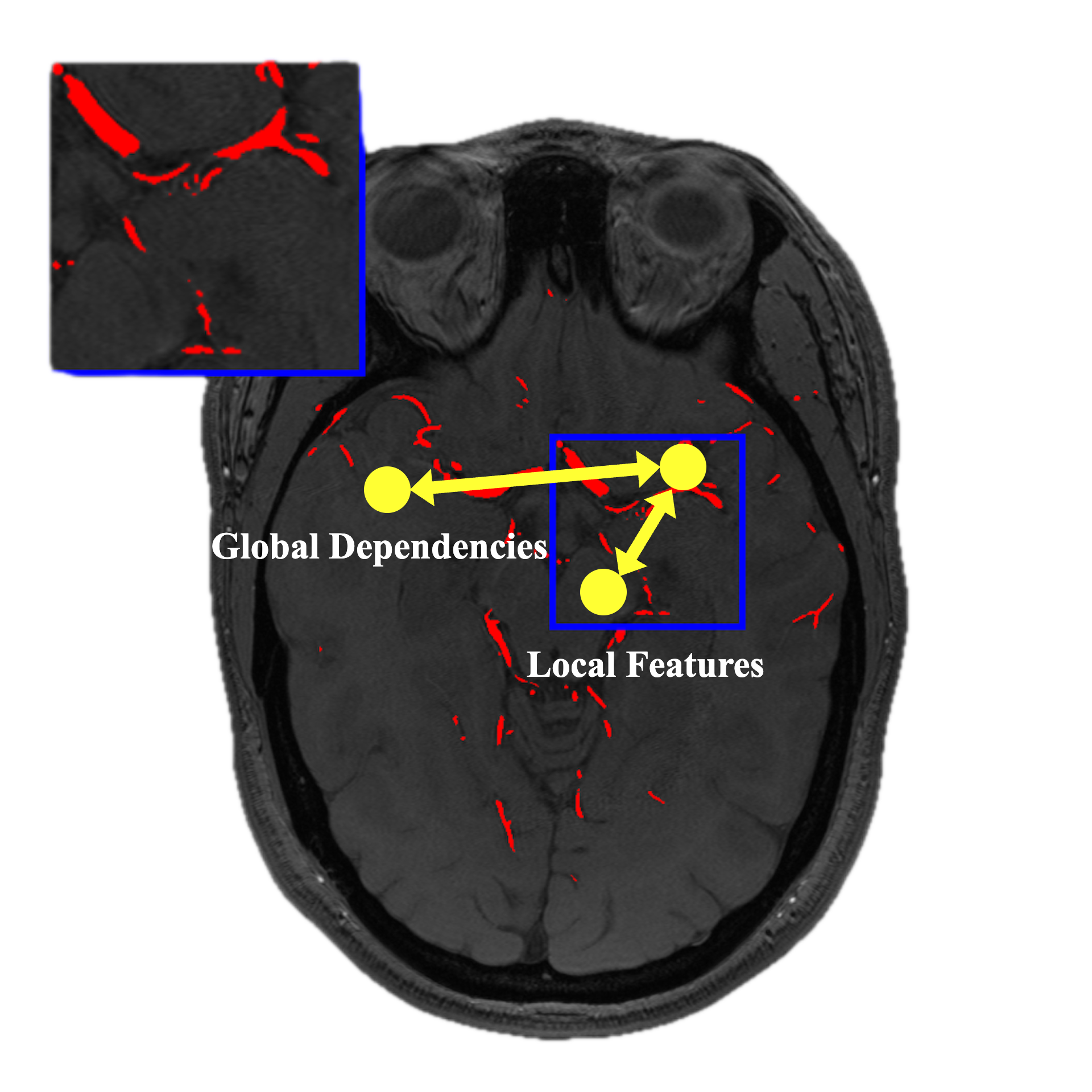}
    \caption{Illustration of local features versus global dependencies in medical vessel segmentation. Blue boxes highlight local feature regions that capture fine-grained vessel details, which are essential for precise boundary delineation. Yellow dots and connecting arrows illustrate long-range spatial relationships between distant vessel segments, enabling the model to understand vessel continuity and overall vascular topology across the entire image. This figure demonstrates the complementary roles of local feature extraction and global dependency modeling using a representative MRA brain scan.}
    \label{fig:CNNT}
\end{figure}

In the field of deep learning, convolutional neural network (CNN), especially those with U-shape encoder-decoder structures \cite{huang2020unet, xiao2020segmentation, isensee2021nnu, xu2021automatic}, have set new benchmarks in various segmentation tasks. In the classic U-Net \cite{ronneberger2015u} architecture, the encoder is used to capture global context by progressively reducing the feature dimensions, while the decoder then enlarges these features back to the original input size for precise pixel or voxel-level segmentation. While CNN-based models \cite{ibtehaz2023acc, li2024lite, xu2023dcsau} excel in representation learning, their ability to capture whole-volume context is constrained by their localized receptive fields due to the limited kernel size \cite{hu2019local}. Moreover, CNN-based approaches face significant computational inefficiency when processing high-resolution 3D medical volumes, as the convolution operations scale poorly with increasing spatial dimensions. Additionally, these models struggle to model long-range spatial dependencies that are crucial for understanding complex anatomical structures, particularly in scenarios where pathological regions may be spatially distant but functionally related. The inherent limitation in capturing global context that spans across different anatomical regions further restricts their effectiveness in comprehensive medical image analysis. As shown in Fig. \ref{fig:CNNT}, CNN-based methods can only extract the feature maps from limited receptive fields, but due to the special topology structure of a human cerebrovascular structure, some global dependencies between vessels may be ignored.

Vision transformer (ViT) \cite{dosovitskiy2020image}, with its self-attention mechanism for global information extraction, gains attention in the realm of 3D medical imaging because of its ability to model long-range dependencies and grasp the global context. Specifically, transformers represent images through a sequence of 1D patch embeddings and mechanisms to learn the relationships among tokens. This enables modeling effective long-range dependencies. Many studies \cite{chen2024transunet, lin2022ds, cao2022swin, chen2024sam, li2024flaws} have demonstrated its superior performance in diverse medical image segmentation tasks. However, the quadratic complexity of non-linear Softmax function in the self-attention mechanism is not efficient in long-sequence imaging computations. Beyond computational complexity, transformer-based approaches face several critical limitations in medical image segmentation. The patch-based tokenization process inherently loses fine-grained spatial details that are essential for precise boundary delineation in medical structures. Furthermore, transformers struggle with multi-scale feature representation, which is particularly challenging when dealing with medical images that contain structures of vastly different sizes, from fine capillaries to major vessels. The substantial memory requirements of transformer models when processing high-resolution 3D medical volumes often necessitate aggressive downsampling or patch-wise processing, potentially compromising the model's ability to maintain spatial coherence across the entire volume. For example, the typically high-resolution 3D volumes produced by the magnetic resonance angiography (MRA) scans result in lengthy 1D sequences, imposing significant computational demands.

To address the limitations of the self-attention mechanism while retaining the advantages of long-sequence modeling, the selective state space model Mamba \cite{gu2023mamba} has been proposed to effectively model long-range dependencies with linear complexity through a novel selection mechanism that enhances the training and inference efficiency. In the field of computer vision, existing pure SSM-based frameworks \cite{ma2024u, ruan2024vm, liu2024swin, xing2024segmamba, wang2024mamba} have revealed significant potential for medical image segmentation and demonstrated lower computation costs compared to ViT-based models \cite{chen2024transunet, he2023h2former, chen2021transunet, lou2025nusegdg, xu2025hrmedseg}. While beneficial, they are difficult to capture local texture information, resulting in performance degradation of segmenting small targets. Therefore, the ideal Mamba-based architectures in medical image segmentation should be able to extract and handle local and global contexts simultaneously.

In our ACM MM Asia 2024 conference paper \cite{chen2024mambavesselnet}, we propose a Hybrid CNN-Mamba architecture that first utilizes Mamba for segmenting 3D cerebrovascular structures, called MambaVesselNet. It contains the hybrid image encoder (Hi-Encoder) that leverages texture-aware layers and vision Mamba blocks to capture low-level semantic features and effectively model long-range dependencies with linear complexity, respectively. After that, MambaVesselNet utilizes the bifocal fusion decoder (BF-Decoder) to comprehensively fuse low-level local texture information and global features by applying skip connection structures for precise segmentation decoding. The main contributions of MambaVesselNet are summarized as follows.
\begin{itemize}
    \item We propose MambaVesselNet which explores the integration of CNN and Mamba for 3D medical segmentation. In particular, this is the first study that explores the integration of CNN and Mamba for segmenting cerebrovascular structures in MRA scans. 
    \item We introduce a Hi-Encoder that contains texture-aware layers to extract local information and vision Mamba layers to model long-range dependencies with linear complexity. Further, we devise the BF-Decoder to combine local texture features and global contexts, leading to precise predictions of segmentation masks.
\end{itemize}

As invited by ACM MM Asia 2024, we extend and further optimize our MambaVesselNet to adapt diverse medical segmentation tasks and propose an updated framework called MambaVesselNet++. By integrating CNN with Mamba modules, our MambaVesselNet++ complementarily takes the respective advantages: CNNs provide detailed local and multi-scale feature representation while Mamba captures global dependencies with linear complexity. Specifically, while MambaVesselNet is limited to 3D cerebrovascular segmentation, MambaVesselNet++ incorporates adaptive convolutions in the Hi-Encoder to support feature extraction from medical images of 2D and 3D, significantly expanding its applicability. Moreover, MambaVesselNet++ adopts a novel multi-branch decoding strategy in the BF-Decoder that enables flexible switching between semantic and instance segmentation tasks, whereas the original framework only supports semantic segmentation. Finally, MambaVesselNet++ demonstrates its adaptability through comprehensive evaluations on five diverse medical modalities (dermoscopy, colonoscopy, fundus, ultrasound, and histopathology), compared to the single-modality focus of MambaVesselNet. Experimental results indicate that our MambaVesselNet++ framework outperforms state-of-the-art medical image segmentation methods, displaying superior performance. The new contributions of MambaVesselNet++ in addition to those of MambaVesselNet are summarized as follows.
\begin{itemize}
    \item This extension proposes the MambaVesselNet++ that further optimizes the prior framework, which leverages adaptive convolution and multi-branch decoding strategy to adapt diverse medical segmentation tasks, including 2D medical semantic and instance segmentation.
    
    \item We validate the effectiveness of our MambaVesselNet++ model on five external public medical modalities, including dermoscopy, colonoscopy, fundus, ultrasound, and histopathology. The proposed model displays superior performance compared to state-of-the-art medical segmentation methods.
\end{itemize}

\section{Related Work}
\textbf{CNN-based Segmentation Models.} Convolution, especially for UNet \cite{ronneberger2015u}, has been widely used in diverse medical segmentation tasks. In 2D imaging modalities, various variants \cite{xu2023dcsau, ibtehaz2023acc, rahman2024emcad} leveraged channel and spatial attention mechanisms to highlight the representation of disease regions. In 3D vascular segmentation, \citet{chen20173d} used innovative CNN architectures for tasks like intracranial artery and vascular boundary segmentation. \citet{tetteh2020deepvesselnet} introduced DeepVesselNet with efficient convolutional filters and a class-balancing loss function. Despite these advancements, challenges like morphological variation and class imbalance in cerebrovascular segmentation in TOF-MRA volumes remained. \citet{yuan2023segmentation} developed a two-stage FCN approach for the segmentation of the aortic vessel tree in CT scans, achieving high accuracy in the MICCAI segmentation challenge. \citet{banerjee2024streamlining} 
proposed a domain-general AI method for the volumetric analysis of cerebrovascular structures across multiple MRA centers using TOF-MRA, employing a multi-task deep CNN with a topology-sensitive loss function to enhance segmentation accuracy. Despite their success in various segmentation tasks, convolutional networks fell short in capturing global context and global spatial dependencies. 

\noindent \textbf{Transformer-based Models.} Transformer-based models made significant strides in the field of medical imaging and computer vision \cite{dosovitskiy2020image, zheng2021rethinking, luomed, wang2025automated, xu2025co, liu2025swin} and medical image analysis \cite{xie2021cotr, xu2024esp, lou2024nusegdg}. To enhance the model's ability to model long-range dependencies, \citet{chen2021transunet} firstly proposed TransUnet, a medical image segmentation framework that combined the global contextual strengths of transformers with the detailed spatial resolution capabilities of U-Net. This concept was further expanded by the introduction of UNETR \cite{hatamizadeh2022unetr}. UNETR utilized the Vision Transformer (ViT) as its encoder to learn the global context and then merged it with a CNN-based decoder through skip connections at various resolutions. Furthermore, SwinUNETR \cite{hatamizadeh2021swin} used the Swin Transformer \cite{liu2021swin} as an encoder. Different from standard transformers, Swin Transformers introduced an innovative hierarchical design that computed self-attention within non-overlapping local windows, and these windows were shifted across layers, allowing the model to capture multi-scale features while maintaining efficiency. However, contemporary 3D medical imaging modalities, such as
magnetic resonance imaging (MRI), often generate highly detailed, multi-volumetric data sets. These high-resolution volumetric scans are translated into lengthy 1D sequences for analysis, thus imposing substantial computational burdens on transformer-based frameworks.

\noindent \textbf{Selective State Space Models.} To address the computational challenges brought by the non-linear softmax function of self-attention, the selective state space model (SSM) \cite{gu2023mamba}, Mamba was proposed to handle long-range dependencies with linear complexity. Mamba enhanced the efficiency of both training and inference by implementing a selection mechanism. In the context of computer vision, Mamba's application was explored in various contexts. U-Mamba \cite{ma2024umamba} designed an SSM-based encoder-decoder framework with the nnUNet style\cite{isensee2021nnu}. Thus, U-Mamba also featured a self-configuring capability that facilitated its autonomous adaptation to varying datasets. Vision Mamba (VMamba) \cite{liu2024vmambavisualstatespace} combined Vision Transformers' global receptive fields with selective state space models' efficiency, introducing the Cross-Scan Module (CSM) for handling non-causal visual data with linear complexity. In terms of 3D medical image segmentation, SegMamba \cite{xing2024segmamba} integrated the Mamba model into Unetr architecture. This novel approach, named Tri-orientated Mamba (ToM), modeled 3D features from multiple directions, enhanced by a Gated Spatial Convolution (GSC) module for spatial feature refinement. However, the above works led to an overemphasis on global contextual information. This overemphasis might result in over-segmentation. For example, non-vessel areas were incorrectly identified as vessels due to the overshadowing of important local features. Conversely, traditional CNN-based models, while effective at capturing local features and spatial details, struggled to model long-range dependencies and global context, leading to under-segmentation where some vessel structures were missed due to their limited receptive fields (as discussed in Section \ref{Dis}). To address these issues, our model retains the advantages of CNN for both the encoder and decoder to leverage its strength in local feature extraction. We incorporate the Mamba block specifically at the bottleneck to effectively model long-range dependencies without compromising the network's ability to accurately localize vessel structures. This strategic placement bridges the gap between high-level feature abstraction and detailed spatial reconstruction, achieving a balance between over-segmentation and under-segmentation.

\section{Methodology}
\begin{figure*}[!t]
    \centering
    \includegraphics[width=1\textwidth]{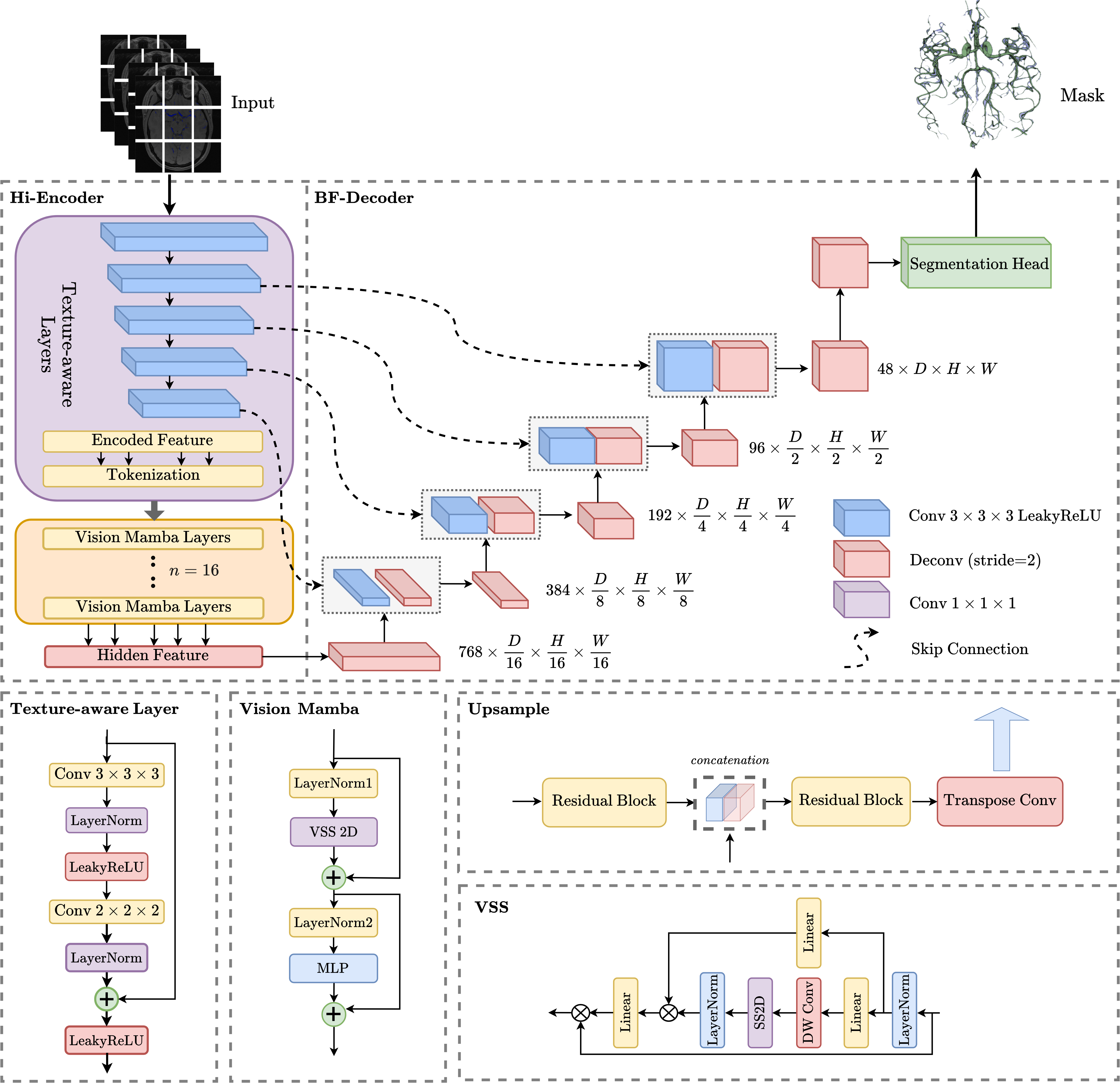}
    \caption{Overview of MambaVesselNet++ architecture. The model takes 2D medical images or 3D patches as the input (channel size $C=4$ for 3D MRA images), and down-samples it through successive texture-aware layers, with each time reducing the spatial resolution by half and doubling the feature channels. After down-sampling and local feature extraction operations, the output feature maps are reshaped into a sequence and fed into the Mamba Blocks ($n=16$). Finally, the BF-Decoder performs the up-sampling operation and restores the reduced spatial dimensions images to their original size, and the skip connection is used to bridge the multi-scale features from the encoder and the decoder.}
    \label{fig:achi}
\end{figure*}

An overview of MambaVesselNet++ architecture is presented in Fig. \ref{fig:achi}. MambaVesselNet++ follows an encoder-decoder framework and is structured into two components: 1) a hybrid image encoder (Hi-Encoder) to capture low-level local texture information and global dependencies from medical imaging, 2) a bifocal fusion decoder (BF-Decoder) to perform a hierarchical decoding workflow and leverages the pixel-level combination of local and global features via skip connections to generate accurate segmentation masks. 

\subsection{Preliminaries}
State Space Models (SSMs) like S4 \cite{gu2021combining} and Mamba can be understood as sophisticated memory systems that process sequential data by maintaining and updating an internal state representation. Conceptually, imagine a system that reads information sequentially (like reading a book page by page) while maintaining a memory of what it has processed so far and uses this accumulated knowledge to make decisions about new incoming information. Mathematically, SSMs modernize classical system theory to handle sequential data by mapping a one-dimensional input \(x(t) \in \mathbb{R}\) to an output \(y(t) \in \mathbb{R}\) through intermediate states \(h(t) \in \mathbb{R}^N.\) According to \cite{gu2023mamba}, this process is framed as:
\begin{equation}
h'(t) = A\cdot h(t) + B\cdot x(t), \quad y(t) = C \cdot h(t),
\end{equation}
where $h(t)$ represents the system's memory or internal state at time $t$, $x(t)$ is the new input information arriving at time $t$, $y(t)$ is the output or decision made based on the current memory state, Matrix $A \in \mathbb{R}^{N \times N}$ controls how to retain or forget past information, Matrix $B \in \mathbb{R}^{N \times 1}$ determines how much new input information is integrated into memory and Matrix $C \in \mathbb{R}^{N \times 1}$ decides how the internal memory state is converted into the final output.

The key innovation of Mamba lies in making the system \textit{selective} rather than using fixed parameters, which dynamically adjusts how it processes information based on the content of the input. When processing a medical image sequence, the selective mechanism can automatically pay more attention to regions with pathological changes, emphasize relevant anatomical features in the output, and adjust the processing speed based on information density. This selective scan mechanism allows Mamba to selectively propagate or filter information along the sequence dimension. Rather than treating all input equally, it can focus computational resources on the most relevant information for the current task, leading to improved performance and computational efficiency. This adaptability is particularly valuable in medical image segmentation, where different regions may require different levels of attention and processing strategies.

\subsection{Hybrid Image Encoder}
The existing Mamba-based U-shape architectures \cite{ma2024umamba, ruan2024vm, liu2024swin} mainly adopt pure Mamaba blocks to model long-range dependencies. Despite the efficiency, they are difficult to capture local features, resulting in missing segmentation details, and degrading performance in tracking small targets. To address this issue, we propose the Hi-Encoder that contains a two-stage layout to deal with the extraction of local semantic information and global contexts. 

\noindent \textbf{Texture-aware Layer.} To build up our Hi-Encoder, we propose a two-stage layout that first employs texture-aware layers to learn local semantic representation efficiently. As depicted in Fig. \ref{fig:achi}, convolutions are used in this module to down-sample the input medical images $\{f_i|i=1,2,\cdots,I\}$, $I$ is the number of layers. Each layer applies two convolutional operations with $3 \times 3$ and $2 \times 2$ kernels, respectively, followed by layer normalization and a LeakyReLU activation function. The convolutional blocks are defined by \( f_i = \text{LeakyReLU}(\text{Norm}_i(W_i \ast f_{i-1} + b_i)) \), where \( f_i \) is the output feature map after the \( i \)-th convolution, \( W_i \) is the \( i \)-th convolutional weight, and \( b_i \) is the \( i \)-th bias. All convolution blocks are designed with a residual style: \( f_{i} \gets \text{LeakyReLU}(f_i)+ f_i \). On this basis, they can effectively group texture information. On the other hand, many U-shape studies \cite{xu2023dcsau, rahman2024emcad, huang2020unet} apply the max-pooling operator to reduce the spatial dimensions. Although it saves parameter costs, max-pooling only retains high-intensity features, which discards low-frequency information. Therefore, we adopt a $2 \times 2$ convolution with stride 2 to capture comprehensive local features. Following each texture-aware layer, the spatial dimensions are reduced by half and the number of feature channels is increased twofold through a down-sampling process. Overall, our texture-aware layer improves the capability of local feature extraction for the Hi-Encoder.

\begin{figure}[!t]
    \centering
    \includegraphics[width=1\linewidth]{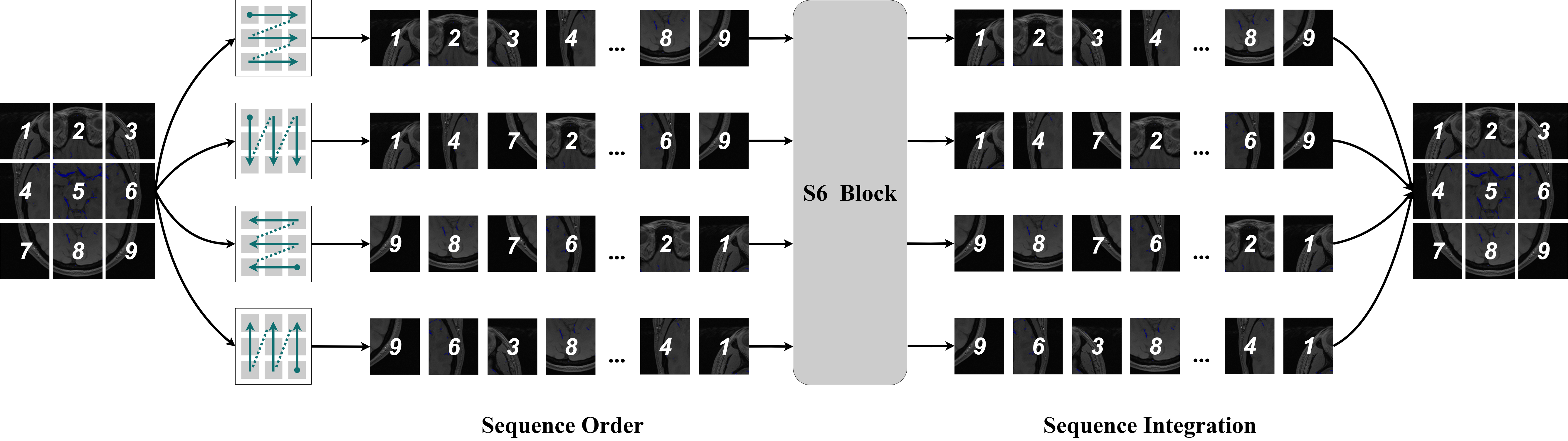}
    \caption{Detailed illustration of the 2D-Selective-Scan (SS2D) mechanism. The SS2D mechanism processes image features through four distinct directional scanning patterns. This multi-directional scanning strategy ensures comprehensive spatial context modeling from all possible directional perspectives, enabling robust feature extraction for medical image segmentation tasks.}
    \label{fig:enc1}
\end{figure}

\noindent \textbf{Vision Mamba Layer.} Many studies \cite{cao2022swin, liu2021swin, chen2024transunet} have demonstrated that modeling feature-level long-range dependencies is more effective in enhancing the diversity of feature representation. Inspired by them, the second stage of the Hi-Encoder adopts the vision Mamaba layer to handle the extraction of feature-level global information with linear complexity, as illustrated in Fig. \ref{fig:achi}. Given the low-level feature map extracted from texture-aware layers, we first flatten it as a set of token sequences, delivering to Mamba blocks. Each block employs an MLP to achieve channel-wise interactions. Then, the updated feature tokens are sent to the visual state space (VSS) block \cite{liu2024vmambavisualstatespace} to grasp global dependencies. As illustrated in Figure \ref{fig:enc1}, SS2D reorganizes feature sequences in four different scan directions. The four sequentially ordered feature sequences are processed through the S6 block, then are reconstructed back to their original spatial arrangement and integrated through element-wise operations to produce the final enhanced feature representation. Specifically, given input feature $z$, the operations of SS2D can be defined by:
\begin{equation}
    z'_n = {\rm order}(z, n), 
\end{equation}
\begin{equation}
    z'_n \gets {\rm S6}(z'_n), 
\end{equation}
\begin{equation}
    z \gets {\rm integrate}(z'_1, z'_2,z'_3,z'_4), 
\end{equation}
where $n \in \{1,2,3,4\}$ stands for four different scanning directions. The whole vision Mamba layer can be written as:
\begin{equation}
\hat{z}^l = \text{Mamba}(\text{LayerNorm}(z^{l-1})) + z^{l-1},
\end{equation}
\begin{equation}
\hat{z}^l_{\text{out}} = \text{MLP}(\text{LayerNorm}(\hat{z}^l)) + \hat{z}^l.
\end{equation}
The Mamba layer processes the normalized input feature map \(z^{l-1}\), producing an intermediate representation \(\hat{z}^l\). This intermediate representation is then normalized again and passed through an MLP, resulting in the final output feature map \(\hat{z}^l_{\text{out}}\). Residual connections are used in both steps to ensure that the original input features (\(z^{l-1}\) and \(\hat{z}^l\)) are added to the corresponding processed features. In this way, our proposed Hi-Encoder captures both low-level local semantic information and global dependencies, improving the diversity of image features.

\subsection{Bifocal Fusion Decoder}
The hierarchical segmentation decoding designed with the U-shape style has revealed high adaptability for predicting masks with different sizes. However, current methods \cite{cao2022swin, xu2023dcsau,liu2024swin,ruan2024vm, ma2024umamba} usually focus on decoding local or global features only, leading to insufficient decoding information and degrading the performance. To tackle this challenge, we devise the Bi-Decoder that leverages the skip connection mechanism to combine the low-level local texture information with long-range dependencies, providing precise segmentation tasks. Specifically, the global feature $G$ extracted from the last vision Mamba layer is regarded as the main decoding target, and is upsampled using a $2 \times 2$ deconvolutional layer with a stride of 2. Then, the updated global feature is fused with the local features from the last texture-aware layer $f_I$ with element-wise addition $\oplus$, followed by a $1\times1$ convolution for feature integration. The above computation $F^{\rm BIF}_I$ can be defined by: 
\begin{equation}
    F^{\rm BIF}_{I} = {\rm Conv_{1\times1}}({\rm DeConv_{2\times2}}(G) \oplus f_I).
\end{equation}
In this way, our Bi-Decoder captures both local texture features and global contexts, achieving bifocal fusion. Next, the feature map in each decoder layer \( k  \in \{1,2\cdots,I-1\}\) is reshaped to dimensions \( \frac{W}{2^k} \times \frac{D}{2^k} \times C_k \), where $C_k=48\times 2^k$. These feature maps are upsampled using the same deconvolutional layer, effectively doubling their resolution. These upsampled outputs are then concatenated with the corresponding outputs $f_k$ from the texture-aware layer via skip connections $\smallfrown$. The combined features are subsequently processed through several $3\times 3$ convolutions to refine the feature representations and generate outputs $F^{\rm BIF}_{k}$ for each stage. The above processes can be formularized as:
\begin{equation}
    F^{\rm BIF}_{k} = {\rm Conv_{3\times3}}({\rm DeConv_{2\times2}}(F^{\rm BIF}_{k+1}) \smallfrown f_k).
\end{equation}
Finally, the segmentation results are produced through a \(1 \times 1\) convolution followed by an activation layer. As a result, the proposed Bi-Decoder effectively takes advantage of local and global dependencies, enhancing the quality of segmentation masks.

\subsection{Optimization Pipeline}
To adapt our MambaVesselNet++ to diverse medical image segmentation tasks, we adopt a classical U-shape style \cite{ronneberger2015u} to construct our model and utilize two key adaptive mechanisms: adaptive convolution for flexible 2D/3D feature extraction and multi-branch decoding for semantic/instance segmentation task switching. Specifically, the adaptive convolution mechanism leverages the difference in the number of dimensions between 2D (i.e., $x\in \mathbb{R}^{B \times C \times H \times W}$) and 3D (i.e., $x\in \mathbb{R}^{B \times C \times H \times W \times D}$) inputs to switch between 2D and 3D convolutional operations. Moreover, the multi-branch decoding strategy contains three decoding heads. For semantic segmentation, our multi-branch decoding only configures one semantic decoding head, which is supervised by Dice and cross-entropy loss functions, as \cite{horst2024cellvit}:
\begin{equation}
\mathcal{L}_{\rm sem}
%(G, Y)
= \lambda_{\text{Dice}} \mathcal{L}_{\rm Dice}
%\left( 1 - \frac{2 G\cdot Y}{G\cdot G+Y\cdot Y}\right)
%\sum\limits_{i=1}^{I} G_{i} Y_{i}}{\sum\limits_{i=1}^{I} G_{i}^2 + \sum\limits_{i=1}^{I} Y_{i}^2} \right) 
+ \lambda_{\text{CE}}\mathcal{L}_{\rm CE}, 
%\left( - \frac{1}{I} \sum\limits_{i=1}^{I} G_{i} \log(Y_{i}) \right),
\end{equation}
where the parameters \( \lambda_{\text{Dice}} \) and \( \lambda_{\text{CE}} \) are the weighting factors for the Dice and cross-entropy components. For instance segmentation, such as in the histopathology modality, the multi-branch decoding strategy follows \cite{graham2019hover, horst2024cellvit} to construct three heads to perform multi-task learning. The first head is responsible for predicting the semantic segmentation map of all nuclei, referred as nuclei prediction (NP), which accurately captures their boundaries and shapes. Meanwhile, the second head produces horizontal and vertical distance maps, known as horizontal-vertical prediction (HV), delivering essential spatial information for accurate localization and detailed delineation. The third head is used to generate nuclei type map (NT). For faster training and better convergence of the network, we employ a combination of different loss functions for each network head. The total loss is defined by:
\begin{equation}
\mathcal{L}_{\text{ins}}=\lambda_{\text{NP}}\mathcal{L}_{\text{NP}}+\lambda_{\text{HV}}\mathcal{L}_{\text{HV}}+\lambda_{\text{NT}}\mathcal{L}_{\text{NT}},
\end{equation}
where $\mathcal{L}_{\text{NP}}$ and $\lambda_{\text{NP}}$ stand for the loss and weight for the NP-head, $\mathcal{L}_{\text{HV}}$ loss  and $\lambda_{\text{HV}}$ for the
HV-head, and $\mathcal{L}_{\text{NT}}$ loss and $\lambda_{\text{NT}}$ for the NT-head. In detail, the losses for each individual head are constructed from the following set of loss functions:
\begin{equation}
\mathcal{L}_{\text{NP}}=\mathcal{L}_{\text{Focal}}+\mathcal{L}_{\text{Dice}},
\end{equation}
\begin{equation}
\mathcal{L}_{\text{HV}}=\mathcal{L}_{\text{MSE}}+\mathcal{L}_{\text{MSGE }},
\end{equation}
\begin{equation}
\mathcal{L}_{\text{NT}}=\mathcal{L}_{\text{Focal}}+\mathcal{L}_{\text{Dice}}+\mathcal{L}_{\text{BCE}}.
\end{equation}
We then apply general morphological algorithms to produce segmentation maps. In summary, our proposed MambaVesselNet++ can be adapted to various medical segmentation tasks. 

\section{Experiments}
This section presents the experimental methodology and results for evaluating the efficacy of our MambaVesselNet++ in medical semantic and instance segmentation tasks, especially for cerebrovascular structures from 3D MRA images.

\subsection{Dataset}
To validate the effectiveness of the proposed MambaVesselNet++, we conduct comprehensive evaluations across diverse medical domains, as shown in Table \ref{tab:datasets}. The details are as follows:
\noindent \textbf{Dermoscopy.} The PH2 dataset \cite{mendoncca2013ph} is a collection of dermoscopic images specifically designed for research on skin lesion analysis, including melanoma detection and segmentation. It contains 200 images with a resolution of $767 \times 576$.

\noindent \textbf{Colonoscopy.} The CVC-ClinicDB dataset \cite{bernal2015wm} is comprised of 612 high-resolution colonoscopy images obtained from 25 different colonoscopy videos. These images are derived from patients undergoing colorectal cancer screening or diagnosis. The dataset is a benchmark for polyp segmentation.

\noindent \textbf{Fundus.} The DRIVE dataset \cite{staal2004ridge} is a widely used benchmark dataset for the evaluation of retinal blood vessel segmentation algorithms. It consists of 20 training and 20 test color fundus photographs of the retina, which were obtained from a diabetic retinopathy screening program in the Netherlands. The images have a size of $565 \times 584$ pixels.

\noindent \textbf{Ultrasound.} The UDIAT dataset \cite{yap2017automated} is obtained from the UDIAT Diagnostic Centre of the Parc Tauli Corporation, Sabadell, Spain, using a Siemens ACUSON scanner for breast lesion segmentation. The dataset contains 163 cases: 109 benign and 54 malignant images, with a single lesion per case. 

\noindent \textbf{Histopathology.} The TNBC dataset \cite{naylor2018segmentation} is a specialized resource designed for advancing computational pathology, particularly in segmenting nuclei instances within histopathological images of aggressive breast cancer subtypes. It includes 50 images with a resolution of $512 \times 512$ pixels, sampled at 40$\times$ magnification.

\noindent \textbf{MRA.} The IXI cerebrovascular dataset \cite{gray2012multi} includes 30 Time-of-Flight Magnetic Resonance Angiography (TOF-MRA) training cases and 15 test cases that are annotated by Chen et al. \cite{chen2022attention} with a resolution of $1024 \times1024 \times 92$, where  Each voxel measuring 0.264 $\times$ 0.264 $\times$ 0.8 mm\(^3\). 

In conclusion, we adopt a usual train-val-test split of 8:1:1 for Dermoscopy, Colonoscopy, Ultrasound, and Histopathology datasets. For Fundus and MRA datasets, we use their official train-val-test sets. 

\begin{table}[!t]
\centering
\small
\caption{Overview of datasets used for evaluation across different medical imaging modalities.}
\label{tab:datasets}
{\scalebox{1}{
\begin{tabular}{lllllll}
\toprule
Modality & Dataset & Images & Resolution & Descriptions \\ \midrule

Dermoscopy & PH2 \cite{mendoncca2013ph} & 200 & $767 \times 576$ & 
Skin lesion segmentation \\ 

Colonoscopy & CVC-ClinicDB \cite{bernal2015wm} & 612 & $384 \times 288$ & 
Polyp segmentation \\

Fundus & DRIVE \cite{staal2004ridge} & 40 & $565 \times 584$ & 
Retinal vessel segmentation \\ 

Ultrasound & UDIAT \cite{yap2017automated} & 163 & Variable & 
Breast lesion segmentation \\

Histopathology & TNBC \cite{naylor2018segmentation} & 50 & $512 \times 512$ & 
Nuclei segmentation \\ 

MRA & IXI \cite{gray2012multi} & 45 & $1024 \times 1024$ & 
Vessel volume segmentation \\ \bottomrule

\end{tabular}%
}}
\end{table}

\subsection{Evaluation Metrics}
\noindent \textbf{Semantic Segmentation Evaluation.} For the quantitative evaluation of our method, we first consider pixel-level errors in medical semantic segmentation tasks, where mean intersection over union (mIoU) and Dice score are standard metrics to measure the similarity between predicted masks and ground truth. The former is derived from the confusion matrix: True Positive (\(TP\)), True Negative (\(TN\)), False Positive (\(FP\)), and False Negative (\(FN\)). When the dataset contains \(N\) samples, mIoU can be defined by:
\begin{equation}
{\text {mIoU}} = \frac{1}{N} 
\sum_{n=1}^{N} {\text {mIoU}}_n,
%\frac{TP_{n}}{TP_{n} + FP_{n} + FN_{n}}.
\end{equation}
where \({\text {mIoU}}_n\)  is the well-known and commonly defined \({\text {mIoU}}\) \cite{xing2024segmamba} of sample $n\in\{1,2,\cdots,N\}$. The Dice score quantifies the degree of overlap between the predicted segmentation and the actual ground truth by:
\begin{equation}
\text{Dice} = \frac{2G \cdot P}{G + P},
\end{equation}
where $G$ is the ground truth and $P$ is the prediction segmentation mask. Additionally, Precision is used to measure the proportion of true positive predictions among all predicted positives and Recall measures the proportion of true positive predictions among all actual positives \cite{yuan2023segmentation}.
\begin{comment}
, as follows:
\begin{equation}
Precision = \frac{TP}{TP + FP},
\end{equation}
\begin{equation}
Recall = \frac{TP}{TP + FN}.
\end{equation}
\end{comment}
The Hausdorff distance (HD) is a metric used to measure the spatial accuracy of predicted boundaries compared to ground truth, as \cite{chen2024transunet}:
\begin{equation}
HD(G, P) = \max_{g \in G} \left( \min_{p \in P} | g - p | \right).
\end{equation}
\noindent \textbf{Instance Segmentation Evaluation.} We further evaluate medical instance segmentation tasks in terms of four metrics \cite{graham2019hover, horst2024cellvit}: panoptic quality (PQ), detection quality (DQ), segmentation quality (SQ) and aggregated Jaccard index (AJI), defined by \cite{horst2024cellvit}:
\begin{equation}
PQ = \underbrace{\frac{|TP|}{|TP| + \frac{1}{2} |FP| + \frac{1}{2} |FN|}}_{\text{Detection Quality (DQ)}} \times \underbrace{\frac{\sum_{(G, P) \in TP} \text{IoU}(G, P)}{|TP|}}_{ \text{Segmentation Quality (SQ)}},
\end{equation}
\begin{equation}
AJI = \frac{\sum_{j=1}^{M} |G_j \cap P_{j}|}{\sum_{j=1}^{M} |G_i \cup P_{j}| + \sum_{k \in K} |P_k|},
\end{equation}
where $M$ denotes the number of instances, $j$ is the instance index and $K$ is the set of predicted instances that do not maximally overlap with any ground truth instance.

\subsection{Implementation Details}
MambaVesselNet++ implementation is based on MONAI $1.2.3$\footnote{\url{https://monai.io/}}, Pytorch and Cuda $11.6$. All experiments are conducted on two NVIDIA A5000 GPUs (48G). The adaptive moment (Adam) optimizer updates the parameters along with the Cosine Annealing Learning Rate Scheduler. The learning rate is initialized at \( 1 \times 10^{-4} \) and annealed down to \( 1 \times 10^{-7} \) over the training. For 3D segmentation, 5000 iterations are used to train all models for convergence. We use a patch size of $64 \times64 \times 64$ and the batch size is set to 2 per GPU. In our implementation, both \(\lambda_{\text{Dice}}\) and \(\lambda_{\text{CE}}\) are set to 1 by default, giving equal weight to both loss components. For 2D segmentation, the number of epoch and batch size is set to 200 and 16. For instance segmentation, the loss coefficients $\lambda_{\rm NP}$, $\lambda_{\rm HV}$ and $\lambda_{\rm NT}$ are set to 1,5 and 1, respectively. The IoU threshold for detecting nuclei instances is set to 0.5, following the standard configuration.

\subsection{Main Results}
\noindent \textbf{Comparison on 2D Medical Semantic Segmentation} To comprehensively illustrate the effectiveness of  
MambaVesselNet++, we compare it with existing state-of-the-art models on 2D medical semantic segmentation tasks, including dermoscopic and polyp segmentation (PH2 \cite{mendoncca2013ph} and CVC-ClinicDB \cite{bernal2015wm} datasets), as well as retinal and ultrasound segmentation (DRIVE \cite{staal2004ridge} and UDIAT \cite{yap2017automated} datasets). The results presented in Table \ref{tab:comparison_sota1} and Table \ref{tab:comparison_sota2} demonstrate that MambaVesselNet++ outperforms all competing models across multiple key evaluation metrics. On the PH2 dataset, MambaVesselNet++ achieves the highest Dice coefficient (0.953) and mIoU (0.911), coupled with the lowest HD (19.50), indicating superior segmentation accuracy and boundary precision. On the CVC-ClinicDB dataset, MambaVesselNet++ continues to lead with a Dice score of 0.911, mIoU of 0.854, and an HD of 13.22, further confirming its robustness in medical image segmentation. In comparison, the other models, such as U-Mamba \cite{ma2024umamba} and TransUNet \cite{chen2024transunet}, exhibit lower Dice and mIoU scores and higher HD values, suggesting inferior segmentation performance in both accuracy and boundary delineation. For retinal segmentation on the DRIVE dataset, MambaVesselNet++ achieves a Dice score of 0.711 and mIoU of 0.552, significantly outperforming TransUNet (Dice: 0.614, mIoU: 0.443). Additionally, on the UDIAT dataset, MambaVesselNet++ achieves the best results with a Dice score of 0.849 and mIoU of 0.763, surpassing TransUNet (Dice: 0.836, mIoU: 0.751) and U-Mamba (Dice: 0.838, mIoU: 0.747). The model also demonstrates lower HD values across most tasks, reflecting superior segmentation boundary precision.

\begin{table}[!t]
\centering
\caption{Quantitative comparison with state-of-the-arts on 2D dermoscopic and polyp semantic segmentation.}
\scalebox{1}{\begin{tabular}{l|ccc|ccc}
\toprule
\multirow{2}{*}{Models} & \multicolumn{3}{c|}{PH2} & \multicolumn{3}{c}{CVC-ClinicDB}\\
\cline{2-7}
& Dice $\uparrow$  & mIoU $\uparrow$ & HD $\downarrow$ & Dice $\uparrow$  & mIoU $\uparrow$ & HD $\downarrow$\\
\midrule
UNext\cite{valanarasu2022unext} & 0.911 & 0.848 & 30.48 & 0.782 & 0.687 & 37.78 \\ 
DCSAU-Net\cite{xu2023dcsau} & 0.919 & 0.859 & 28.44 & 0.781 & 0.695 & 29.68 \\ 
TransUNet\cite{chen2024transunet} & 0.936 & 0.886 & 31.18 & 0.899 & 0.837 & 17.97 \\ 
U-Mamba\cite{ma2024umamba}  & 0.904 & 0.936 & 77.96 & 0.896 & 0.842 & 30.79\\
Swin-UMamba\cite{liu2024swin} & 0.941 & 0.892 & 25.72 & 0.904 & 0.849 & 21.26 \\
MambaVesselNet++ & \textbf{0.953} & \textbf{0.911} & \textbf{19.50} & \textbf{0.911} & \textbf{0.854} & \textbf{13.22} \\ 
\bottomrule
\end{tabular}}
\label{tab:comparison_sota1}
\end{table}

\begin{table}[!t]
\centering
\caption{Quantitative comparison with state-of-the-arts on 2D retinal and ultrasound semantic segmentation.}
\scalebox{1}{\begin{tabular}{l|ccc|ccc}
\toprule
\multirow{2}{*}{Models} & \multicolumn{3}{c|}{DRIVE} & \multicolumn{3}{c}{UDIAT}\\
\cline{2-7}
& Dice $\uparrow$  & mIoU $\uparrow$ & HD $\downarrow$ & Dice $\uparrow$  & mIoU $\uparrow$ & HD $\downarrow$\\
\midrule
UNext\cite{valanarasu2022unext} & 0.531 & 0.382 & 30.14 & 0.667 & 0.539 & 45.94 \\
DCSAU-Net\cite{xu2023dcsau} & 0.512 & 0.345 & 29.89 & 0.677 & 0.574 & 34.69 \\
TransUNet\cite{chen2024transunet} & 0.614 & 0.443 & 21.68 & 0.836 & 0.751 & \textbf{18.68} \\
U-Mamba\cite{ma2024umamba} & 0.449 & 0.301 & 53.01 & 0.761 & 0.695 & 28.08 \\
Swin-UMamba\cite{liu2024swin} & 0.601 & 0.429 & 19.41 & 0.838 & 0.747 & 18.83 \\
MambaVesselNet++ & \textbf{0.711} & \textbf{0.552} & \textbf{18.12} & \textbf{0.849} & \textbf{0.763} & 23.01 \\
\bottomrule
\end{tabular}}
\label{tab:comparison_sota2}
\end{table}

\begin{table}[!t]
\centering
\caption{Quantitative comparison with state-of-the-arts on 2D nuclei instance segmentation.}
\begin{tabular}{l|cccc}
\toprule
Models & AJI $\uparrow$  & PQ $\uparrow$ & DQ $\uparrow$ & SQ $\uparrow$\\
\midrule
Hover-Net\cite{huang2020unet} & 0.503 & 0.442 & 0.620 & 0.711\\ 
StarDIST\cite{isensee2021nnu} & 0.457 & 0.405 & 0.592 & 0.685\\ 
CellPose\cite{hatamizadeh2022unetr} & 0.511 & 0.468 & 0.647 & 0.723\\ 
CPP-Net\cite{liu2021swin} & 0.482 & 0.446 & 0.629 & 0.708\\ 
CellViT\cite{xing2024segmamba} & 0.517 & \textbf{0.481} & 0.647 & \textbf{0.746}\\ 
MambaVesselNet++ & \textbf{0.534} & 0.479 & \textbf{0.657} & 0.729\\ 
\bottomrule
\end{tabular}
\label{tab:comparison_sota3}
\end{table}

\begin{table}[!t]
\centering
\caption{Quantitative comparison with state-of-the-arts on 3D cerebrovascular segmentation.}
\begin{tabular}{l|ccc|cc}
\toprule
\multirow{2}{*}{Models} & \multicolumn{3}{c|}{Performance Metrics} & \multicolumn{2}{c}{Training Cost} \\
\cline{2-4} \cline{5-6}
& Precision $\uparrow$ & Recall $\uparrow$ & Dice $\uparrow$ & Time (min/epoch) $\downarrow$ & Memory (GB) $\downarrow$ \\
\midrule
Unet3D\cite{huang2020unet} & 0.863 & 0.805 & 0.831 & 4.5 & 8.2 \\ 
nnUNet\cite{isensee2021nnu} & 0.856 & 0.845 & 0.849 & 2.8 & 7.6 \\ 
UNETR\cite{hatamizadeh2022unetr} & 0.835 & 0.842 & 0.836 & 3.1 & 9.4 \\ 
SwinUNETR\cite{liu2021swin} & 0.845 & \textbf{0.874} & 0.857 & 3.8 & 9.8 \\ 
SegMamba\cite{xing2024segmamba} & 0.863 & 0.870 & 0.864 & 2.3 & 7.3 \\ 
MambaVesselNet++ & \textbf{0.889} & 0.859 & \textbf{0.870} & \textbf{2.1} & \textbf{6.9} \\ 
\bottomrule
\end{tabular}
\label{tab:comparison_sota4}
\end{table}

\begin{table}[!t]
  \centering
  \setlength\tabcolsep{7pt}
  \caption{Ablation study of MambaVesselNet++ on the 3D IXI dataset.}
  {\scalebox{0.98}{
  \begin{tabular}{l|ccccc}
  \toprule
  Methods & Precision $\uparrow$ & Recall $\uparrow$ & Dice $\uparrow$ \\
  \midrule
  Baseline  & 0.835 & 0.842 & 0.836   \\
  + Texture-aware Layer & 0.842 & 0.837 & 0.840 \\
  + Vision Mamba Layer & 0.855 & 0.841 & 0.851 \\
  + Hi-Encoder & 0.867 & 0.851 & 0.859 \\
  + Hi-Encoder \& BF-Decoder (Ours) &  \textbf{0.889} & \textbf{0.859} & \textbf{0.870} \\

  \bottomrule
  \end{tabular}}}
  \label{tab:ab}
\end{table}

\noindent \textbf{Comparison on 2D Nuclei Instance Segmentation}
MambaVesselNet++ is compared against several state-of-the-art models on the 2D nuclei instance segmentation task, as shown in Table \ref{tab:comparison_sota2}. The models evaluated include Hover-Net \cite{graham2019hover}, StarDIST \cite{schmidt2018cell}, CellPose \cite{stringer2021cellpose}, CPP-Net \cite{chen2023cpp}, and CellViT \cite{horst2024cellvit}. The comparison is based on four key metrics: Average Jaccard Index (AJI), Panoptic Quality (PQ), Detection Quality (DQ), and Segmentation Quality (SQ). MambaVesselNet++ demonstrates competitive performance across all metrics. It achieves an AJI of 0.534, a PQ of 0.479, a DQ of 0.657, and an SQ of 0.729. Among the models compared, MambaVesselNet++ outperforms most of them in terms of Segmentation Quality (SQ), where it achieves the highest value of 0.746. Additionally, MambaVesselNet++ surpasses Hover-Net (SQ: 0.711), StarDIST (SQ: 0.685), and CPP-Net (SQ: 0.708) in this regard. In terms of AJI, MambaVesselNet++ achieves the highest score of 0.534, followed closely by Cellvit with an AJI of 0.517. For PQ, MambaVesselNet++ achieves a score of 0.481, which is slightly higher than the scores of models like Hover-Net (PQ: 0.442) and StarDIST (PQ: 0.405), indicating its advantage in accurately identifying and segmenting individual nuclei.

\noindent \textbf{Comparison on 3D Vessel Volume Segmentation} In this evaluation, we benchmark MambaVesselNet++ against five state-of-the-art segmentation models, assessing their performance on cerebrovascular segmentation. These include two CNN-based approaches, UNet3D \cite{huang2020unet} and nnUNet \cite{isensee2021nnu}, and two transformer-based methods, UNETR \cite{hatamizadeh2022unetr} and SwinUNETR \cite{liu2021swin}. Additionally, we compare it with another Mamba-based model, SegMamba \cite{xing2024segmamba} specialized in medical image segmentation. All models are trained and tested under the same data augmentation settings, and the public implementations of these models from MONAI are used to generate the best results. As shown in Table \ref{tab:comparison_sota4}, MambaVesselNet++ outperforms all other models in terms of Dice score and Precision. Specifically, MambaVesselNet++ achieves the highest Dice score of $0.870$ and the highest Precision of $0.889$. Although its Recall of $0.859$ is slightly lower than that of SwinUNETR ($0.874$) and SegMamba ($0.870$), the higher Precision of MambaVesselNet++ contributes to a better overall Dice score. This balance between Precision and Recall is crucial, as it reflects the model's ability to accurately identify vessel structures while minimizing false positives. MambaVesselNet++ increases the Dice score by approximately $3.9\%$ compared to UNet3D, $2.1\%$ compared to nnUNet, and $0.6\%$ compared to the previous Mamba model SegMamba. The superior performance of MambaVesselNet++ underscores the effectiveness of our proposed Hybrid CNN-Mamba architecture. Additionally, the Swin-Transformer-based model, SwinUNETR, achieves the highest Recall but with lower Precision, resulting in a lower Dice score compared to MambaVesselNet++.

\subsection{Ablation Study}
To further investigate the effectiveness of the Hi-Encoder and BF-Decoder modules of our proposed MambaVesselNet++, an ablation study is conducted on the 3D IXI dataset \cite{gray2012multi}, as shown in Table \ref{tab:ab}. The study compares the performance of several model variants, starting with the original UNETR \cite{hatamizadeh2022unetr} as our baseline model and progressively adding specific modules, including the texture-aware layer, the vision Mamba layer, the Hi-Encoder, and the BF-Decoder. The baseline model achieves a Precision of 0.835, Recall of 0.842, and Dice score of 0.836. When a texture-aware layer is added, precision improves slightly to 0.842, while recall decreases marginally to 0.837, resulting in a small increase in Dice score (0.840). Incorporating the vision Mamba layer leads to a more substantial improvement in precision (0.855) and a slight decrease in recall (0.841), resulting in an increased Dice score of 0.851. The addition of the full Hi-Encoder further boosts precision to 0.867 and recall to 0.851, contributing to a Dice score of 0.859. Finally, the combination of the Hi-Encoder and BF-Decoder achieves the best performance, with precision reaching 0.889, recall at 0.859, and Dice at 0.870. These comparisons demonstrate the effectiveness of each module in our proposed MambaVesselNet++. The final configuration, with the Hi-Encoder and BF-Decoder, shows significant improvements in all metrics, particularly in precision and Dice score, confirming the efficiency of their combination in enhancing the overall segmentation accuracy of MambaVesselNet++.

\begin{figure}[!t]
    \centering
    \includegraphics[width=1\linewidth]{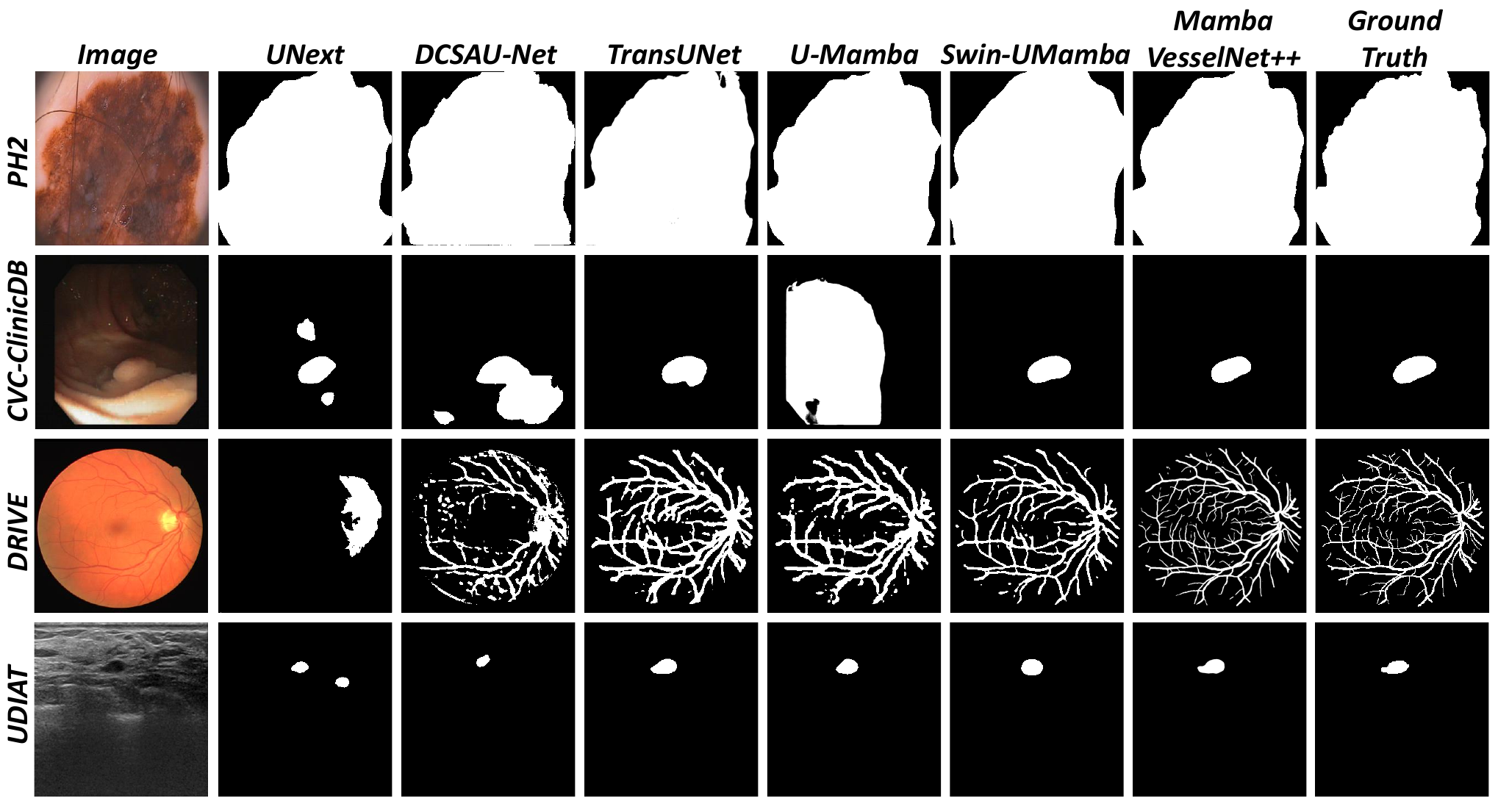}
    \caption{2D medical semantic comparison of different baselines on PH2, CVC-ClinicDB, DRIVE and UDIAT datasets. Our MambaVesselNet++ displays the best results, segmenting accurate target regions and boundaries.}
    \label{fig:sem}
\end{figure}

\begin{figure}[!t]
    \centering
    \includegraphics[width=1\linewidth]{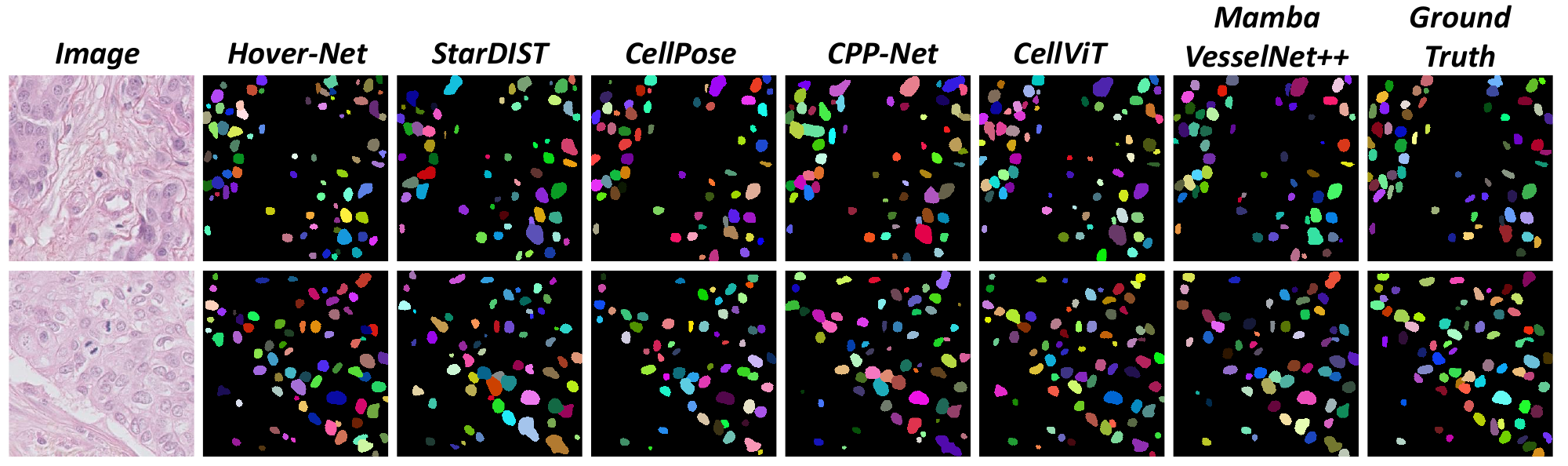}
    \caption{2D nuclei instance comparison of different baselines on the TNBC dataset. Our MambaVesselNet++ exhibits the best results, segmenting more correct nuclei with fewer false positives.}
    \label{fig:ins}
\end{figure}

\begin{figure}[!t]
    \centering
    \includegraphics[width=1\linewidth]{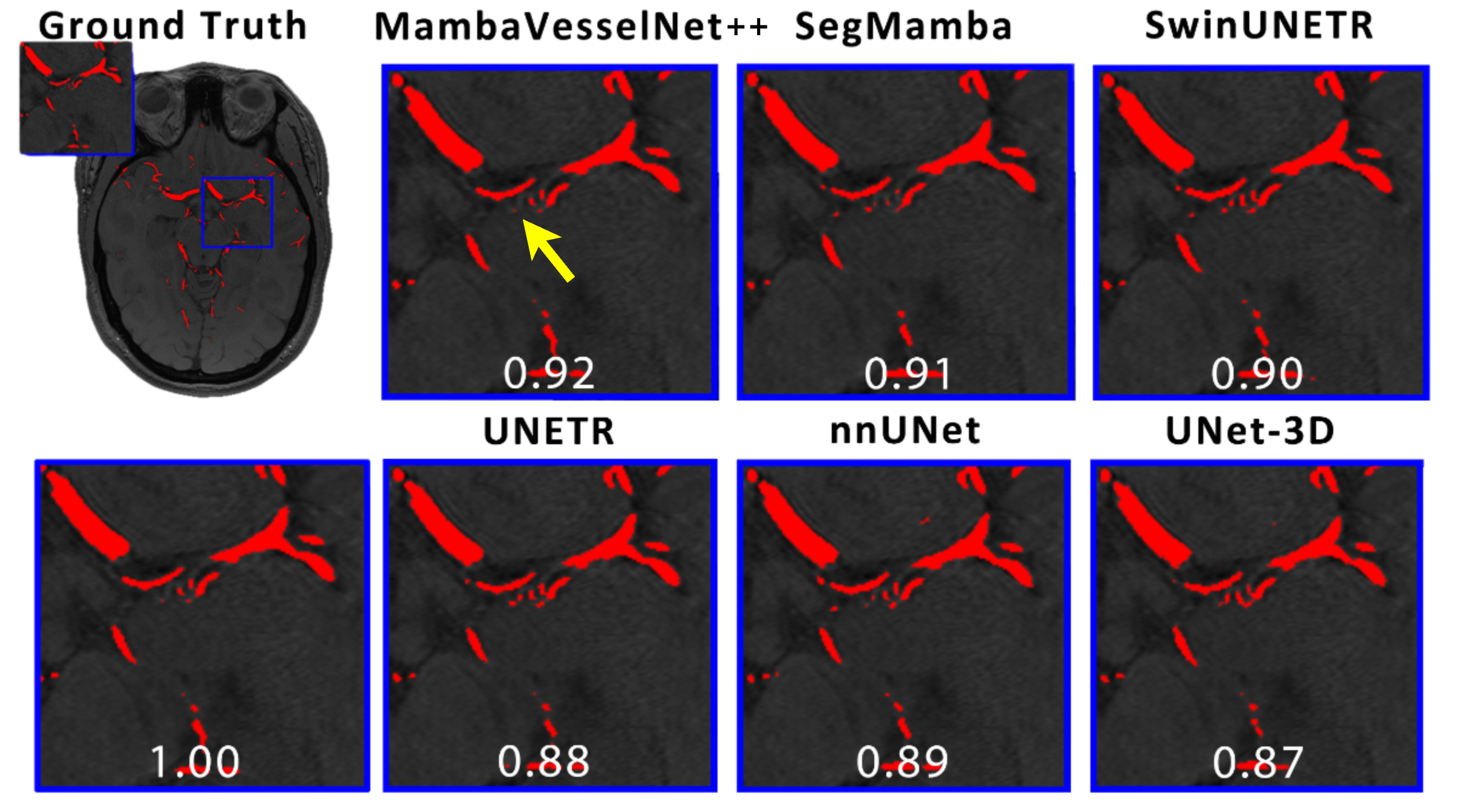}
    \caption{2D qualitative comparison of different baselines on the IXI dataset. The first image shows the ground truth, with a blue box zooming into a part of the region. It is followed by MambaVesselNet++ (Dice score 0.92), SegMamba (Dice score 0.91), SwinUNETR (Dice score 0.90), UNETR (Dice score 0.88), nnUNet (Dice score 0.89), and UNet-3D (Dice score 0.87). Each image highlights the segmentation quality in comparison to the ground truth. The Dice score shown represents the segmentation performance on this specific image.}
    \label{fig:R1}
\end{figure}

\begin{figure}[!t]
    \centering
    \includegraphics[width=1\linewidth]{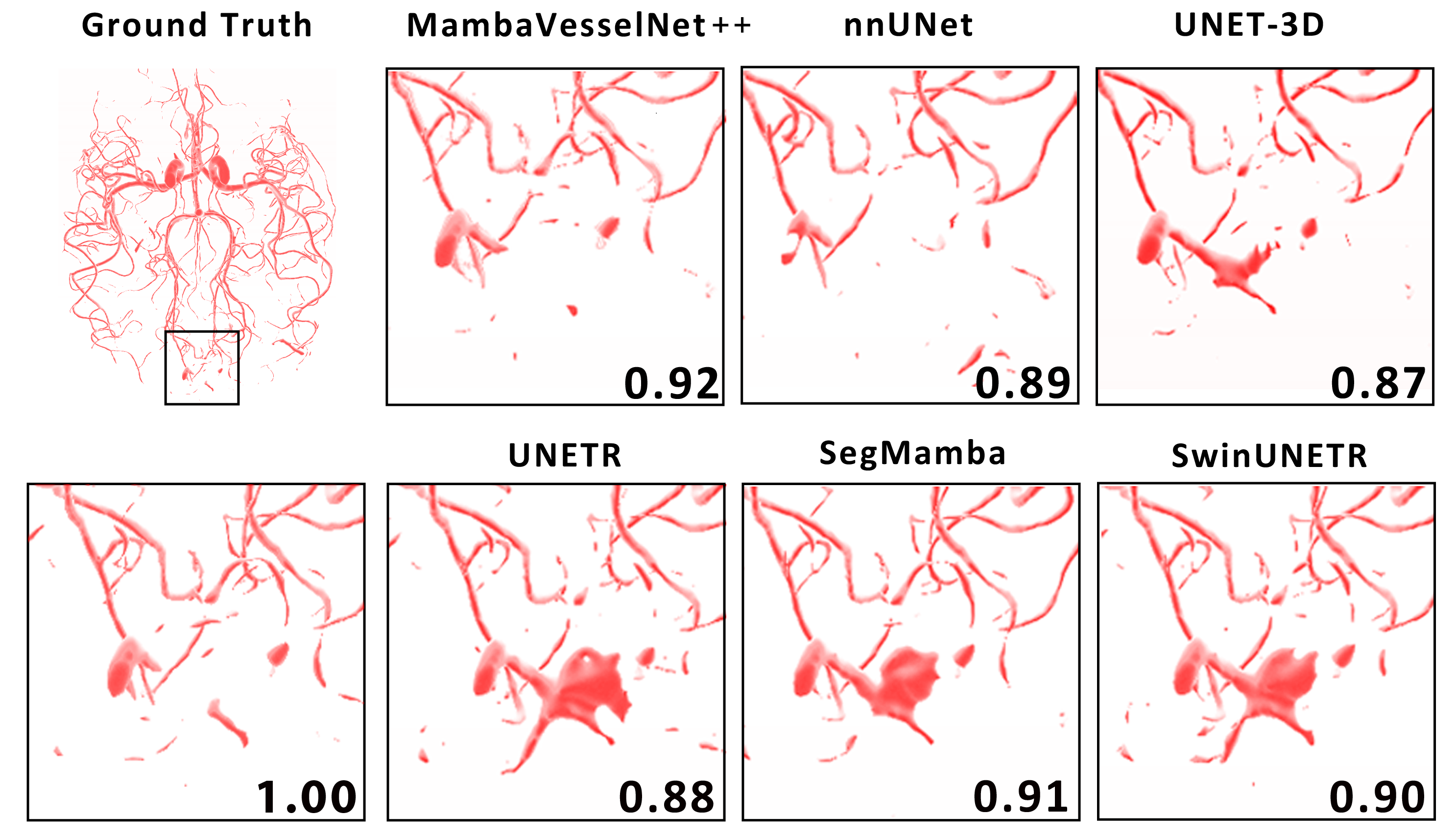}
    \caption{
   This figure shows a 3D qualitative comparison of segmentation results on the IXI dataset. The first image displays the ground truth segmentation. The subsequent images show the results from MambaVesselNet++ (Dice score 0.92), nnUNet (Dice score 0.89), UNET-3D (Dice score 0.87), UNETR (Dice score 0.88), SegMamba (Dice score 0.91), and SwinUNETR (Dice score 0.90). Each model's segmentation quality is compared to the ground truth. The Dice score shown represents the segmentation performance on this specific image.}
    \label{fig:R2}
\end{figure}

\subsection{Comparison on Domain Generalization}

We further validate the generalization capability of our MambaVesselNet++ across different medical imaging domains. To assess the model's ability to generalize to unseen domains, we conduct domain transfer experiments on four distinct medical segmentation tasks, training on one dataset and testing on another dataset from the same anatomical region but with different imaging characteristics. As shown in Table~\ref{tab:comparison_gen1}, we evaluate the domain generalization performance on dermoscopic and colonoscopic modalities. For dermoscopic lesion segmentation, we train models on the PH2 dataset and test on the ISIC2018 dataset \cite{tschandl2018ham10000, codella2019skin}. Our MambaVesselNet++ achieves superior performance with a Dice score of 83.1\%, mIoU of 75.3\%, and Hausdorff Distance of 25.17, significantly outperforming all baseline methods. Notably, MambaVesselNet++ demonstrates substantial improvements over the second-best performing method Swin-UMamba, with gains of 1.7\% in Dice score, 2.2\% in mIoU, and 29.4\% reduction in Hausdorff Distance. For polyp segmentation, training on CVC-ClinicDB and testing on ColonDB \cite{vazquez2017benchmark}, our method achieves the best results across all metrics with 71.2\% Dice, 62.5\% mIoU, and 33.90 HD, demonstrating consistent generalization capability across different endoscopic imaging conditions.

Table~\ref{tab:comparison_gen2} presents the domain shift evaluation results for the retinal vessel and ultrasound breast lesion segmentation tasks. For retinal vessel segmentation, we train on the DRIVE dataset and evaluate on the STARE dataset~\cite{hoover2003locating}. Our MambaVesselNet++ achieves remarkable performance with 60.4\% Dice score and 49.7\% mIoU, substantially outperforming competing methods. The improvement is particularly significant compared to U-Mamba, showing 22.6\% and 24.6\% relative improvements in Dice and mIoU respectively, highlighting the effectiveness of our selective state-space mechanism in handling cross-domain vessel structures. For ultrasound breast lesion segmentation, training on UDIAT and testing on BUSI \cite{al2020dataset}, our method consistently delivers the best performance with 67.3\% Dice, 58.9\% mIoU, and 29.16 HD. These cross-domain results further confirm the practical value of our MambaVesselNet++ framework for real-world clinical deployment, where models often encounter imaging data with different characteristics from the training distribution. The substantial performance margins demonstrate that our approach maintains reliability when applied to unseen medical imaging domains.

\begin{table}[!t]
\centering
\caption{Comparison with state-of-the-arts on unseen dermoscopic and polyp domains.}
\scalebox{1}{\begin{tabular}{l|ccc|ccc}
\toprule
\multirow{2}{*}{Models} & \multicolumn{3}{c|}{PH2 $\rightarrow$ ISIC2018} & \multicolumn{3}{c}{CVC-ClinicDB $\rightarrow$ ColonDB}\\
\cline{2-7}
& Dice $\uparrow$  & mIoU $\uparrow$ & HD $\downarrow$ & Dice $\uparrow$  & mIoU $\uparrow$ & HD $\downarrow$\\
\midrule
UNext\cite{valanarasu2022unext} & 0.783 & 0.692 & 42.15 & 0.654 & 0.531 & 48.92 \\ 
DCSAU-Net\cite{xu2023dcsau} & 0.791 & 0.701 & 39.87 & 0.647 & 0.524 & 46.33 \\ 
TransUNet\cite{chen2024transunet} & 0.806 & 0.723 & 38.42 & 0.681 & 0.567 & 41.28 \\ 
U-Mamba\cite{ma2024umamba} & 0.774 & 0.685 & 45.23 & 0.658 & 0.548 & 47.85 \\ 
Swin-UMamba\cite{liu2024swin} & 0.814 & 0.731 & 35.64 & 0.689 & 0.575 & 39.71 \\ 
MambaVesselNet++ & \textbf{0.831} & \textbf{0.753} & \textbf{25.17} & \textbf{0.712} & \textbf{0.625} & \textbf{33.90} \\ 
\bottomrule
\end{tabular}}
\label{tab:comparison_gen1}
\end{table}

\begin{table}[!t]
\centering
\caption{Comparison with state-of-the-arts on unseen retinal and ultrasound domains.}
\scalebox{1}{\begin{tabular}{l|ccc|ccc}
\toprule
\multirow{2}{*}{Models} & \multicolumn{3}{c|}{DRIVE $\rightarrow$ STARE} & \multicolumn{3}{c}{UDIAT $\rightarrow$ BUSI}\\
\cline{2-7}
& Dice $\uparrow$  & mIoU $\uparrow$ & HD $\downarrow$ & Dice $\uparrow$  & mIoU $\uparrow$ & HD $\downarrow$\\
\midrule
UNext\cite{valanarasu2022unext} & 0.447 & 0.318 & 41.76 & 0.592 & 0.463 & 52.18 \\ 
DCSAU-Net\cite{xu2023dcsau} & 0.433 & 0.299 & 43.52 & 0.604 & 0.481 & 49.83 \\ 
TransUNet\cite{chen2024transunet} & 0.521 & 0.374 & 40.47 & 0.618 & 0.498 & 32.45 \\ 
U-Mamba\cite{ma2024umamba} & 0.378 & 0.251 & 58.64 & 0.643 & 0.537 & 41.27 \\ 
Swin-UMamba\cite{liu2024swin} & 0.568 & 0.438 & 38.19 & 0.661 & 0.565 & 31.98 \\ 
MambaVesselNet++ & \textbf{0.604} & \textbf{0.497} & \textbf{36.45} & \textbf{0.673} & \textbf{0.589} & \textbf{29.16} \\ 
\bottomrule
\end{tabular}}
\label{tab:comparison_gen2}
\end{table}

\section{Discussion}\label{Dis}
The proposed MambaVesselNet++ reveals superior performance over both CNN-based, transformer-based, and existing Mamba-based frameworks on diverse medical segmentation tasks. By taking advantage of the vision Mamba at the bottleneck of the network, MambaVesselNet++ can effectively model long-range dependencies. This approach has improved segmentation accuracy without incurring additional computational burdens. To further conduct qualitative comparisons, we first visualize 2D medical semantic segmentation results in Fig. \ref{fig:sem}. It can be observed that our MambaVesselNet++ performs better in localizing target boundaries and can provide more correct segmentation details, such as tiny retina blood vessels. Moreover, we visualize nuclei instance segmentation map in Fig. \ref{fig:ins}. It illustrates that our MambaVesselNet++ is able to segment nuclei more accurately with fewer false positives. For the 3D MRA modality, 2D cerebrovascular segmentation visualization is presented in Fig. \ref{fig:R1}. It shows that the Mamba-based methods, MambaVesselNet++ and SegMamba, identify finer cerebral vessels more accurately compared to traditional CNN-based approaches. In the central region of the magnified images, varying degrees of false positives are visible in all baseline methods but our benchmark MambaVesselNet++ exhibits fewer false positive markings and produces the segmentation closest to ground truth. In addition, as presented in Fig. \ref{fig:R2}, the qualitative evaluation reveals distinct differences in segmentation performance among various models. CNN-based models, such as nnUNet and UNet3D, tend to exhibit under-segmentation, missing some vessel structures due to their limited receptive fields and inability to capture long-range dependencies. However, they produce fewer false positives because they focus on local features and are less likely to misclassify non-vessel regions. In contrast, transformer-based models (UNETR and SwinUNETR) and the previous Mamba-based model (SegMamba), which utilize transformers or Mamba components as the backbone or encoder, excel at capturing global contextual information. This strength allows them to recognize long-range dependencies between vessel structures, leading to more complete segmentation. However, the overemphasis on global features can overshadow critical local details, resulting in over-segmentation where non-vessel areas are incorrectly segmented as vessels. Our proposed MambaVesselNet++ addresses these challenges by maintaining a traditional CNN-based design in both the encoder and decoder to ensure precise local feature extraction and spatial localization. By introducing the Mamba block exclusively at the bottleneck, we enable the model to capture essential global dependencies without letting them dominate the entire network. This design choice allows us to harness the benefits of both local and global information, effectively balancing high-level feature abstraction with detailed spatial reconstruction. Consequently, MambaVesselNet++ achieves superior segmentation performance by mitigating the over-segmentation seen in transformer or Mamba backbone models and the under-segmentation observed in purely CNN-based models, closely matching the ground truth.

Furthermore, the selective state-space mechanism in MambaVesselNet++ provides inherent interpretability advantages for clinical applications. The selective parameters ($B$ and $C$) in our framework offer information on \textit{ on what} the model focuses on during segmentation. For example, in cerebrovascular segmentation, high selective $B$ values indicate regions where the model integrates more input information, typically corresponding to vessel bifurcations or pathological areas that require careful attention. Similarly, selective $C$ values reveal the features that the model considers the most relevant for final segmentation decisions, helping clinicians verify whether the model focuses on clinically meaningful and anatomical structures. In addition, understanding model confidence is crucial for decision-making. Our MambaVesselNet++ framework provides interpretable confidence measures through the selective mechanism's dynamic parameter adaptation. When the selective parameters exhibit high variance across spatial locations, they often indicate the regions of uncertainty where the model is less confident about the segmentation boundaries. Meanwhile, stable selective parameters suggest high-confidence predictions. This uncertainty quantification allows clinicians to identify the case requiring a manual review or an additional imaging, thereby enhancing patient safety and diagnostic accuracy. Overall, the interpretability features of MambaVesselNet++ position it as an effective clinical decision support tool rather than a replacement for clinical expertise. By providing both accurate segmentation and transparent reasoning through selective attention visualization, the model enables clinicians to: 1) quickly identify areas requiring detailed examination; 2) verify AI predictions against clinical knowledge; 3) understand model confidence levels for risk assessment; and 4) integrate AI insights into comprehensive diagnostic workflows. This interpretable approach addresses a critical barrier to AI adoption in healthcare by providing the transparency necessary for clinical trust and regulatory approval. The combination of high performance and interpretability makes MambaVesselNet++ particularly suitable for clinical applications.

\section{Conclusion}
In this paper, extended and updated from MambaVesselNet, we have proposed MambaVesselNet++, a novel Hybrid CNN-Mamba architecture for medical image segmentation that effectively combines local feature extraction with global dependency modeling. The MambaVesselNet++ framework is comprised of a Hi-Encoder and a BF-Decoder, designed with the U-shape style. Specifically, we first devise the Hi-Encoder that integrates texture-aware layers for capturing local semantic features with vision Mamba blocks for modeling long-range dependencies with linear complexity. We then introduce the BF-Decoder that leverages skip connections to combine local and global information, enabling precise segmentation mask generation. Through extensive experiments across six publicly available medical datasets, including dermoscopic, colonoscopy, retinal, ultrasound, and histopathology, and 3D MRA modalities, we have demonstrated that MambaVesselNet++ outperforms existing state-of-the-art methods, including CNN-based, transformer-based and current Mamba-based models. In particular, our MambaVesselNet++ has significantly reduced computational costs in 3D medical segmentation tasks, which is clinical-friendly in practical scenarios.

\begin{acks}
This work is partially supported by the Yongjiang Technology Innovation Project (2022A-097-G), Zhejiang Department of Transportation General Research and Development Project (2024039), and National Natural Science Foundation of China Grant (UNNC: B0166).
\end{acks}

%%
%% The next two lines define the bibliography style to be used, and
%% the bibliography file.
\bibliographystyle{ACM-Reference-Format}
\bibliography{sample-base}

\end{document}